\documentclass{article}

\usepackage{arxiv}

\usepackage[utf8]{inputenc} % allow utf-8 input
\usepackage[T1]{fontenc}    % use 8-bit T1 fonts
\usepackage{color}

\usepackage[colorlinks=true,linkcolor=black,citecolor=blue]{hyperref}%

\usepackage{url}            % simple URL typesetting
\usepackage{booktabs}       % professional-quality tables
\usepackage{amsfonts}       % blackboard math symbols
\usepackage{nicefrac}       % compact symbols for 1/2, etc.
\usepackage{microtype}      % microtypography
\usepackage{lipsum}
\usepackage{enumitem}
\usepackage{natbib}
\usepackage{amsmath,amsfonts,bm}

\def\Secref#1{Section~\ref{#1}}
\def\eqref#1{Eq.~(\ref{#1})}
\def\1{\bm{1}}

\def\vh{{\bm{h}}}

\def\vp{{\bm{p}}}

\def\vw{{\bm{w}}}
\def\vx{{\bm{x}}}

\def\vz{{\bm{z}}}

\DeclareMathAlphabet{\mathsfit}{\encodingdefault}{\sfdefault}{m}{sl}
\SetMathAlphabet{\mathsfit}{bold}{\encodingdefault}{\sfdefault}{bx}{n}

\def\gD{{\mathcal{D}}}

\def\gF{{\mathcal{F}}}

\def\gL{{\mathcal{L}}}

\def\gN{{\mathcal{N}}}
\def\gO{{\mathcal{O}}}

\def\gT{{\mathcal{T}}}

\def\gW{{\mathcal{W}}}
\def\gX{{\mathcal{X}}}

\newcommand{\R}{\mathbb{R}}

\DeclareMathOperator*{\argmax}{arg\,max}

\usepackage{amsmath,amsfonts,bm,amsthm}
\newtheorem{theorem}{Theorem}[section]

\usepackage{algorithmic}
\usepackage{algorithm}

\usepackage{amsfonts}
\theoremstyle{definition}

\usepackage{Definitions}
\usepackage{graphicx}
\DeclareMathOperator*{\esssup}{ess\,sup}

\title{Review: Ordinary Differential Equations\\
For Deep Learning}

\author{
Xinshi Chen
}

\begin{document}
\maketitle

\tableofcontents

\section{Introduction}
Deep neural network is one of the most popular models in many machine learning tasks. To better understand and improve the behavior of neural network, a recent line of works bridged the connection between ordinary differential equations (ODEs) and deep neural networks (DNNs). The connections are made in two folds:
\begin{itemize}
    \item View DNN as ODE discretization;
    \item View the training of DNN as solving an optimal control problem.
\end{itemize}
The former connection motivates people either to design neural architectures based on ODE discretization schemes or to replace DNN by a continuous model charaterized by ODEs. Several works demonstrated distinct advantages of using continuous model (ODE) instead of traditional DNN in some specific applications~\citep{grathwohl2018ffjord, zhang2018monge,chen2019particle}.
The latter connection is inspiring. Based on Pontryagin's maximum principle, which is popular in the optimal control literature, some developed new optimization methods for training neural networks and some developed algorithms to train the infinite-deep continuous model with low memory-cost.

This paper is organized as follows: In Section~\ref{sec:archi}, the relation between neural architecture and ODE discretization is introduced. Some architectures are not motivated by ODE, but they are latter found to be associated with some specific discretization schemes. Some architectures are designed based on ODE discretization and expected to achieve some special properties. Section~\ref{sec:conti-model} formulate the optimization problem where a traditional neural network is replaced by a continuous model (ODE). The formulated optimization problem is an optimal control problem. Therefore, two different types of controls will also be discussed in this section. In Section~\ref{sec:opt}, we will discuss how we can utilize the optimization methods that are popular in optimal control literature to help the training of machine learning problems. Finally, two applications of using continuous model will be shown in Section~\ref{sec:app1} and~\ref{sec:app2} to demonstrate some of its advantages over traditional neural networks. 

\section{ODE for Neural Architecture}\label{sec:archi}

 A line of works~\citep{liao2016bridging,li2017maximum,chen2018neural,lu2018beyond,ruthotto2018deep,weinan2019mean,chang2018antisymmetricrnn} has mentioned the relation between residual networks and discretization schemes of ODEs. More specifically, for an ODE in the form
\begin{equation}\label{eq:ode}
d X(t)/dt = f(X,W),
\end{equation}
its approximation using forward Euler discretization scheme is
\begin{equation}\label{eq:euler-disc}
X_{t+1} = X_t + hf(X_t,W_t),
\end{equation}
which can be regarded as a generalization of residual networks~\citep{liao2016bridging,chen2018neural,lu2018beyond,ruthotto2018deep,chang2018antisymmetricrnn}. \citet{lu2018beyond} showed some previous effective networks including PolyNet~\citep{zhang2017polynet} and FractalNet~\citep{larsson2017fractalnet} which are not motivated by ODE can also be interpreted as different discretizations of differential equations. After that, \citet{lu2018beyond} proposed LM-ResNet which corresponds to linear multi-step discretization scheme, \citet{chang2018antisymmetricrnn} proposed AntisymmetricRNN by utilizing stability analysis of ODE, \citet{ruthotto2018deep} proposed 2ndOrderCNN and HamiltonianCNN. A summary is given in table~\ref{tab:architecture}.

\begin{table}[h!]
    \centering
        \caption{Different deep networks and their associated ODE discretization schemes.}
    \begin{tabular}{c|l|c}
        \hline
         architecture  & formula &  discretization \\
         \hline 
         ResNet~\citep{he2016deep} & $X_{t+1} = X_t + f(X_t)$ & Forward Euler \\
         & {\scriptsize refer to only one previous point}\\
         \hline
         PolyNet~\citep{zhang2017polynet} & $X_{t+1}=X_t + f(X_t) + f(f(X_t)) $&  Backward Euler \\
          & $\quad \quad ~\approx (I-f)^{-1}\rbr{X_t}$ \\
         & {\scriptsize implict scheme, stabler than forward Euler}\\
         \hline
         FractalNet~\citep{larsson2017fractalnet} & $f_{k+1}(X_t) = \sbr{\rbr{f_{k}\circ f_k}(X_t)}\bigoplus \sbr{\text{conv}(X_t)}$ & Runge-Kutta \\
         &  {\scriptsize take some intermediate steps to obtain a higher order method}\\
         \hline
         LM-ResNet~\citep{lu2018beyond} & $X_{t+1} = (1-h_t)X_t + h_t X_{t-1} + f(X_t)$ & Linear multistep\\
             &{\scriptsize use information from the previous step to calculate the next value}\\
         \hline
         2ndOrderCNN~\citep{ruthotto2018deep} & $X_{t+1}= 2X_t-X_{t-1}+h_t^2 f(X_t) $ & Leapfrog method \\
         &{\scriptsize  one type of multistep methods} \\
         \hline
                  AntisymmetricRNN~\citep{chang2018antisymmetricrnn} & $H_t = H_{t-1} + \varepsilon \sigma\rbr{(W-W^\top)H_{t-1}+VX_t + b } $ & Forward Euler\\
         &{\scriptsize Jacobian matrix has eigenvalues of zero real part}\\
         \hline
         RevNet~\citep{gomez2017reversible} & $Y_t = X_t + f(X_{t+1})$,  $Y_{t+1} = X_{t+1} + g(Y_t)$ \\ 
         HamiltonCNN~\citep{ruthotto2018deep} & $Y_{t+1} = Y_t + f(X_t)$, $X_{t+1} = X_t - g(Y_{t+1})$& a system of ODE\\
    \hline
    \end{tabular}
    \label{tab:architecture}
\end{table}

Despite from these interesting connections, a question arises: why are these architectures expected to be better? Although there are experimental evidences in these works which reveal some benefits of using these architectures, there is no clear explanation of the reason. The connection to ODE discretization also does not help much on our understanding, because numerical discretization schemes are typically designed to improve the accuracy of approximating the solution of ODEs (e.g. Runge-Kutta and multi-step methods), while a neural network is typically not targeted at solving certain ODEs.

Among these works, there are two interesting angles. First, for the RevNet and HamiltonianCNN which are reversible, it allows us to avoid storage of intermediate network states, thus achieving higher memory efficiency.  This is particularly important for very deep networks where memory limitation can hinder training. Second, \citet{haber2017stable,chang2018antisymmetricrnn} introduced an intriguing idea of utilizing the stability theory of ODE. A stable network can be more robust to the perturbation (or noise) of the inputs. Unfortunately, the architecture proposed by \citet{haber2017stable,chang2018antisymmetricrnn} are based on a set of conditions that in general can not ensure stability. Nevertheless, this ideas is still inspiring.

\section{ODE for Continuous Modeling}\label{sec:conti-model}
Instead of using discretization of ODE to design neural architecture, one can directly use a continuous model. Many machine learning tasks are learning some function mappings $F:\R^d \rightarrow \R $ from data. The function mapping $F$ is usually a neural network with unknown parameters. Now, one can consider the mapping from $\vx$ to $F(\vx)$ as an evolution from initial state $X(0)=\vx$ to the finial state $X(T)$, where the dynamics of $X(t)$ for $0\leq t\leq T$ can be modeled by a differential equation, $dX/dt = f$. Learning the mapping $F$ becomes learning the flow velocity $f:\Omega \rightarrow \R^d$ which determines continuous propagation of $X(t)$. 

With the traditional neural network model replaced by a continuous ODE, a supervised learning problem can be formulated as 
% The continuous formulation can help us bridge machine learning problems with optimal control problems with. Inspired by existing tools that are popular in optimal control literature, many new ....
% \subsection{Continuous Formulation}
% Supervised learning is the machine learning task of learning a function $F:\R^d \rightarrow \R $ based on example pairs $\{(\vx^i,y^i)\}_{i=1}^N$. One can consider the mapping from $\vx^i$ to $F(\vx^i)$ as an evolution from initial state $X(0)=\vx^i$ to the finial state $X(T)$, where the dynamics of $X(t)$ can be modeled by a differential equation. With the traditional neural network model replaced by a continuous ODE, a supervised learning problem can be formulated as 
\begin{align}
   \min_{W}~ & L \left(X(T)\right) + \int_{0}^{T} R(W)\,dt \label{eq:loss} \\
  \text{s.t.}~\dfrac{dX(t)}{dt} &= f(X(t), W, t),\quad X(0)=\vx_0,\quad 0\leq t\leq T, \label{eq:ode-constraint}
\end{align}
where $L(\cdot)$ is the estimation loss function, $R$ is the running cost or regularity, and $W$ controls the propagation of $X(t)$. The control $W$ is similar to the parameters/weights of a neural network. Next, we will discuss two questions:
\begin{enumerate}
    \item How to solve the optimal $W$?
    \item What is the benefit of using a continuous model? Why not use traditional neural networks?
\end{enumerate}
For the second question, several works demonstrated distinct advantages of using continuous model instead of traditional DNN in some specific applications~\citep{grathwohl2018ffjord, zhang2018monge,chen2019particle}, which are summarized in Section~\ref{sec:app1} and Section~\ref{sec:app2}. For the first question, the ODE constraint optimization problem in~\eqref{eq:loss} and \eqref{eq:ode-constraint} can be regarded as a standard optimal control problem, where the unknown $W$ is referred to {\bf control} in optimal control literature. Optimal control has a rich history, so there are many tools that can be utilized for machine learning problems. \citet{li2017maximum,weinan2019mean} first introduced the idea of regarding a supervised learning problem as an optimal control problem. They consider training the parameters of a neural network as solving an optimal control $W$ from a collection of admissible controls.

\begin{figure}[h!]
    \centering
    \includegraphics[width=0.42\textwidth]{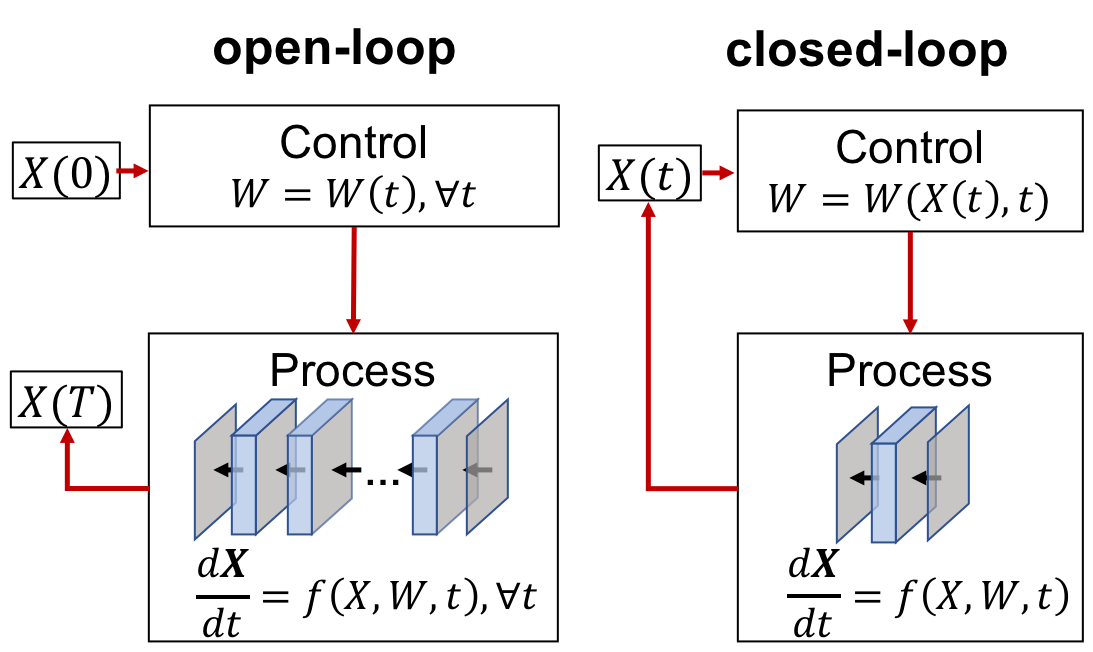}
    \caption{{\it Left}: An open-loop control makes a sequence of decisions when the initial state is observed. {\it Right}: A closed-loop control makes decisions based on the most updated information.}
    \label{fig:closed-loop}
\end{figure}

In optimal control literature, the control $W$ can be modeled in two forms. One is the feed-back form $W:\rbr{X(t),t}\mapsto W(X(t),t)\in\gW$, which is called {\bf closed-loop control}. The other form is $W:t\mapsto W(t)\in\gW$ (or $W: \rbr{X(0),t}\mapsto W(X(0),t)\in\gW$), which is called {\bf open-loop control}. Closed-loop control is in general a stronger charaterization of the solution of the control problem. It can actively adapt to the encountered states $X(t)$. In contrast, open-loop control will determine the whole path $W(t)$ for all $t\in[0,T]$ at the initial time $t=0$, so it is not adapted to $X(t)$. See Figure~\ref{fig:closed-loop} for a visual explanation. 

Closed-loop control is useful for {\it stochastic} optimal control problems (i.e. the flow $f$ is stochastic), because there exists uncertainty and $W$ can adapted to the most updated information. However, the closed-loop control $W(X(t),t)$ is hard to solve. Typically, one needs to consider the {\bf value function}
$
V(t, \vx) := \inf_{W\in\gW} \int_{t}^T R(W)\,dt + L(X(T)),~\forall (t,\vx)\in [0,T]\times \gX,
$
%where $X(t)=\vx$ and the associated ODE constraint $dX= f(X,W,t)dt$ is satisfied. The value function 
which satisfies the {\bf Hamilton-Jacobi-Bellman} (HJB) {\it equantion}:
\begin{align}\label{eq:hjb}
    &-V_t + \sup_{W\in \gW}H(t,\vx,-V_{\vx},W)=0,\quad \forall (t, \vx)\in [0,T)\times \gX
    \end{align}
and $V(T,\vx) = L(\vx)$, where $H:[0,T]\times \gX \times \R^d \times \gW \rightarrow \R$ 
is called the {\bf Hamiltonian} and defined as
\begin{align}
    H(t,\vx, \vp, W):= \vp\cdot f(\vx, W, t) - R(W).\label{eq:hamilton}
\end{align}
Typically, an optimal closed-loop control is obtained by solving the HJB equation, which is in general hard to solve numerically. Although the solution of HJB equation is well-studied, the algorithms are ususally not scalable. Thus it is hard to apply them for high-dimensional machine learning problems. Conversely, many works are trying to use deep learning tools to solve high-dimensional differential equations~\citep{hutzenthaler2019proof,grohs2018proof,han2018solving,freno2019machine,weinan2017deep} and also high-dimensional optimal control problems~\citep{jentzen2018proof,reisinger2019rectified}. 
In fact, reinforcement learning is solving stochastic optimal control problems. It generalizes and extends ideas from optimal control to non-traditional control problems.

One possibility to make the closed-loop control easier to solve is to use a biased model. For example, one can parameterize the closed-loop control as $W=\phi(X(t), t; \theta)$ where $\phi$ is a model (e.g. a neural network) and $\theta\in\Theta$ is its learnable parameters. Then the weights output from $\phi$ will be adaptive to the current state $X(t)$. This is similar to HyperNetwork: a widely known approach of using one network to generate the weights for another network.

To conclude, closed-loop control is hard to solve in general, and existing algorithms may not be useful for machine learning problems. Meanwhile, for deterministic problems, an open-loop control is good enough. In fact, in a deterministic system, the optimal closed-loop control and the optimal open-loop control will give the same control law and thus the same optimality~\citep{dreyfus1964some}. Therefore, for the remaining of this paper, we will focus on open-loop control and continue to investigate the two questions posted above.

\section{Optimization Methods}\label{sec:opt}
In this section, we will discuss existing methods of solving the optimal open-loop control, and how to utilize it for machine learning problems.

\subsection{Pontryagin's Maximum Principle}
For solving the open-loop control, Pontryagin's Maximum Principle (PMP)~\citep{pontryagin1962mathematical,boltyanskii1960theory} gives a set of first order necessary conditions for the optimal pair $(X^*, W^*)$. In PMP, there is an important concept: the {\bf adjoint process} $P(t)=-\frac{\partial L(X^*(T))}{\partial X^*(t)}$. It can represent the gradient of loss with respect to the state $X^*(t)$. Its evolution is determined by the ODE in~\eqref{eq:H-eq-2}. The following is the statement of PMP for deterministic problem.
\begin{theorem}{\bf \textrm(Pontryagin's Maximum Principle)} Let $W^*(\cdot)\in\gW$ be the optimal control and $X^*$ the optimal controlled state trajectory. Suppose $\esssup_{t\in [0,T]}\|W^*(t)\|_{\infty}<\infty$. Then there exists an adjoint process $P:[0,T]\rightarrow\R^d$ such that $\forall t\in [0,T]$,
\begin{align}
&\dot{X}^*(t)= \nabla_{\vp}H(t, X^*(t), P(t), W^*(t)),\quad X^*(0) = \vx_0, \label{eq:H-eq-1} \\
&\dot{P}(t)=- \nabla_{\vx} H(t, X^*(t), P(t), W^*(t)),\quad P(T) = -\nabla_{\vx}L(X^*(T)),\label{eq:H-eq-2}\\
&W^*(t) =\argmax_{W\in \gW} H(t, X^*(t), P(t), W),\quad \forall t\in [0,T], \label{eq:pmp-eq-3}
\end{align}
where $H$ is the Hamiltonian defined in~\eqref{eq:hamilton}.
\end{theorem}

\eqref{eq:pmp-eq-3} can not be implemented without discretization. There is also a discrete version of PMP, where the state $X$ is updated iteratively by $X_{k+1} = g(X_k,W_k,k)$. The corresponding discrete PMP states the following conditions:
\begin{align}
&X^*_{t+1}= g(X^*_t,W^*_t,t),\quad X^*_0 = \vx^i, \\
&P_t=- \nabla_{\vx} H(t, X^*_t, P_{t+1}, W^*_t),\quad P_T = -\nabla_{\vx}L(X^*_T),\\
& W^*_t= \argmax_{W\in \gW} H(t, X^*_t, P_{t+1}, W),\quad \forall t=0,\ldots,T-1,\label{eq:dispmp-eq-3}
\end{align}
where $H$ is the discrete Hamiltonian 
\begin{align}
    H(t,\vx, \vp, W):=\vp \cdot g(\vx,W,t) - \delta R(W).
\end{align}
Hence, by solving PMP, one can obtain the optimal open-loop control.
There are many ways of solving PMP. However, to solve large scale problems, Method of successive approximation (MSA) is a more favorable algorithm. 
\begin{algorithm}[ht!]
\caption{discrete MSA}
\label{alg:MSA}
\begin{algorithmic}[1]
\STATE Initialize $W^0$.
\FOR{$k=0$ to M}
    \STATE Given $X_0=\vx_0$, compute $X_{t+1} = g(X_t,W^{(k)}_t,t)$, for $t=0$ to $T-1$;
    \STATE Given $P_T=-\nabla_{\vx}L(X_T) $, compute $P_t=- \nabla_{\vx} H(t, X_t, P_{t+1}, W_t^{(k)})$ for $t=T-1$ to $0$;
    \STATE Set $W^{(k+1)}_t= \argmax_{W'\in \gW} H(t, X_t, P_{t+1},  W')$ for each $t=0,\ldots,T-1$.\label{msa:line:max}
\ENDFOR
\end{algorithmic}
\end{algorithm}

There is an interesting fact that if one replaces line~\ref{msa:line:max} in algorithm~\ref{alg:MSA} by one gradient step $W^{(k+1)}_t\gets W^{(k+1)}_t + \nabla_W  H(t, X_t, P_{t+1},  W^{(k)}_t)$, then it is equivalent to gradient descent~\citep{li2017maximum}. 

To conclude, PMP provides another optimization method other than stochastic gradient descent. However, what is the benefit of using PMP instead of SGD? (1) The first obvious advantage is that PMP does not require gradient with respect to $W$. Therefore, it can be applied to train discrete variables. There are also experiments showing that very sparse weights are obtained while using PMP to train discrete neural networks~\citep{li2018optimal}. (2) Thanks to the rich history of optimal control, PMP has the advantages that rigorous error estimates and convergence results can be established, which allows it easier to modify and improve the algorithm. For example,~\citet{li2017maximum} modified the Hamiltonian and showed a favorable initial convergence rate per iteration.

\subsection{Compute Gradient for Continuous Model}
\begin{figure}[h]
    \centering
    \includegraphics[width=0.7\textwidth]{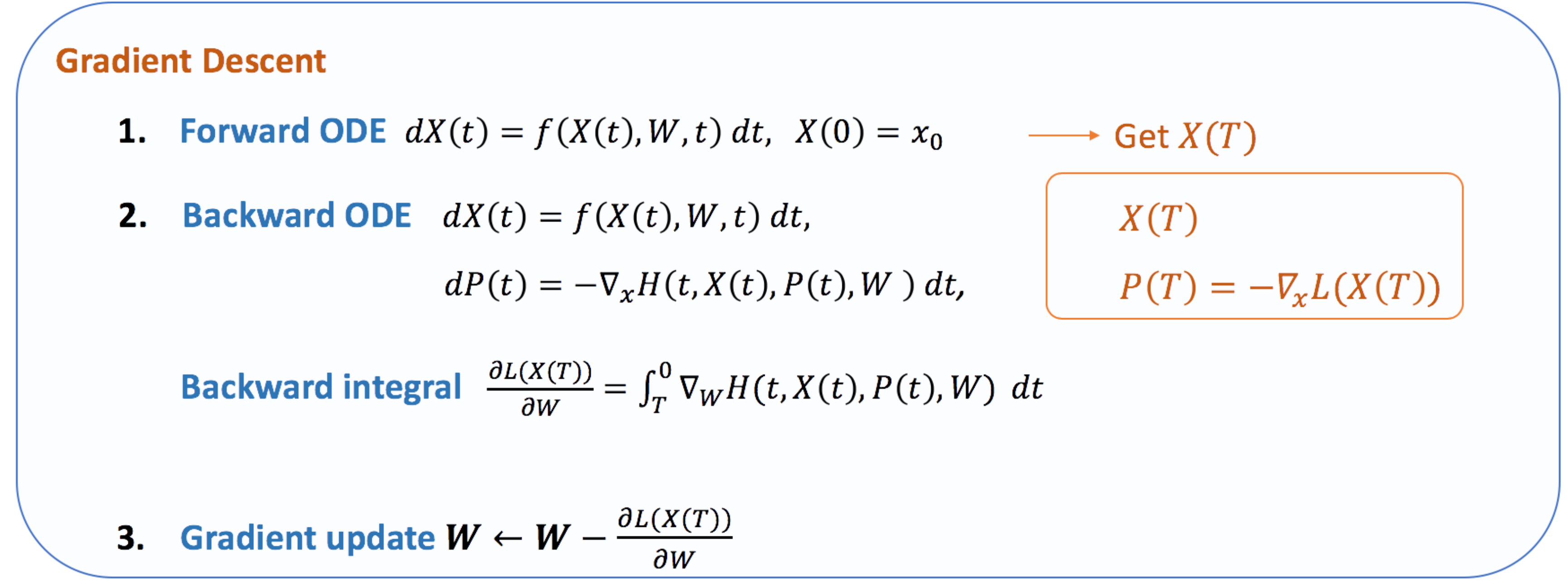}
    \caption{Use adjoint process to compute gradient with low memory cost.}
    \label{fig:adjoint}
\end{figure}

For a high-dimensional {\it continuous} model, it is expensive to use PMP to solve $W$ because of the final step in~\eqref{eq:pmp-eq-3}. However, one can simplify the model by setting a time-invariant control $W(t)\equiv W$ and make use of the adjoint process to compute the gradient. \citet{chen2018neural} proposed this idea and provided an algorithm which only requires $O(1)$ memory. A summary is given in Figure~\ref{fig:adjoint}.
\citet{chen2018neural} also showed by experiments that compared to ResNet, this continuous model can achieve comparable performance with a much smaller number of parameters $W(t)\equiv W$. In the next two sections, we will discuss two more applications to show the advantages of continuous model in some important machine learning problems.

\section{Application: Generative Model}\label{sec:app1}

Suppose the data points $\cbr{\vx^i}_{i=1}^N$ are collected from a particular distribution $p_{x}$. The goal of generative modeling is to estimate the distribution $p_{x}$ given a collection of data points. A common way is to start from a simple distribution $Z\sim p_z$ (e.g. standard Gaussian $\gN(0,1)$), and then transform the variable $Z$ using a function mapping $X= \phi_{\theta}(Z)$, where $\phi_{\theta}$ can be a neural network with parameters $\theta$~\citep{dinh2014nice,kingma2018glow,dinh2016density}. $\theta$ will be learned using the data points. However, to compute the transformed density, one needs to compute 
\begin{align}
    \log p_x(\vx) = \log p_z(\vz) - \log \det \left|\frac{\partial \phi_{\theta}(\vz)}{\partial\vz} \right|.
\end{align}
The determinant of Jacobian requires $\gO(d^3)$ computation, so existing methods have to restrict the architecture of $\phi_{\theta}$ to make this term tractable. 

\citet{zhang2018monge} and \citet{grathwohl2018ffjord} used continuous ODE to replace the function mapping $\phi_{\theta}$. That is, sampling from an initial distribution $X(0)\sim p_z$, through a continuous propagation determined by the ODE $\frac{dX}{dt}=f(X,W,t)$, the solution $X(T)$ is the sample from the transformed distribution. The advantage of this modeling is that the change of density also follows an ODE
\begin{align}
    \frac{d \log p_x(\vx,t)}{dt} = -\nabla_x \cdot f, 
\end{align}
where the computation of $-\nabla_x \cdot f$ is $\gO(d^2)$. Through a stochastic estimator,~\citet{grathwohl2018ffjord} reduced this computation cost to $\gO(d)$. By applying theorems from optimal transport,~\citet{zhang2018monge} proposed to model the flow $f$ as the gradient of a neural network $f:=\nabla \phi_{\theta}$. 

\section{Research Work: Particle Flow Bayes' Rule}\label{sec:app2}
\subsection{Introduction and Related Work}
Bayesian inference is a core machine learning problem. In many data analysis tasks, it is important to estimate unknown quantities $\vx \in \R^d$ from observations $\gO_m:=\{o_1,\cdots,o_m\}$. Given prior knowledge $\pi(\vx)$ and likelihood functions $p(o_t|\vx)$, the essence of Bayesian inference is to compute the posterior $p(\vx|\gO_m)\propto \pi(\vx)\prod_{t=1}^{m} p(o_t|\vx)$ by Bayes' rule. For many nontrivial models, the prior might not be conjugate to the likelihood, making the posterior not in a closed form. Therefore, computing the posterior often results in intractable integration and poses significant challenges. Typically, one resorts to approximate inference methods such as sampling (e.g., MCMC)~\citep{andrieu2003introduction} or variational inference~\citep{wainwright2008graphical}. 

In many real problems, observations arrive sequentially online, and Bayesian inference needs be performed recursively, 
\begin{equation}\label{eq:recursive-Bayes}
     \overbrace{p(\vx|\gO_{m+1})}^\text{updated posterior}~\propto \overbrace{p(\vx|\gO_{m})}^\text{current posterior}\overbrace{p(o_{m+1}|\vx)}^\text{likelihood}. 
\end{equation}
That is the estimation of $p(\vx|\gO_{m+1})$ should be computed based on the estimation of $p(\vx|\gO_{m})$ obtained from the last iteration and the presence of the new observation $o_{m+1}$. 
It therefore requires algorithms which allow for efficient online inference. In this case, 
both standard MCMC and variational inference become inefficient, since the former requires a complete scan of the dataset in each iteration, and the latter requires solving an optimization for every new observation. Thus, sequential Monte Carlo (SMC)~\citep{doucet2001introduction,balakrishnan2006one} or stochastic approximations, such as stochastic gradient Langevin dynamics~\citep{welling2011bayesian} and stochastic variational inference~\citep{hoffman2013stochastic}, are developed to improve the efficiency. However, SMC suffers from the path degeneracy problems in high dimensions~\citep{daum2003curse}, and rejuvenation steps are designed but may violate the online sequential update  requirement~\citep{canini2009online,chopin2013smc2}. 
Stochastic approximation methods are prescribed algorithms that cannot exploit the structure of the problem for further improvement.  

\begin{figure*}[t!]
    \vspace{-1mm}
    \centering
    \includegraphics[width=0.99\textwidth]{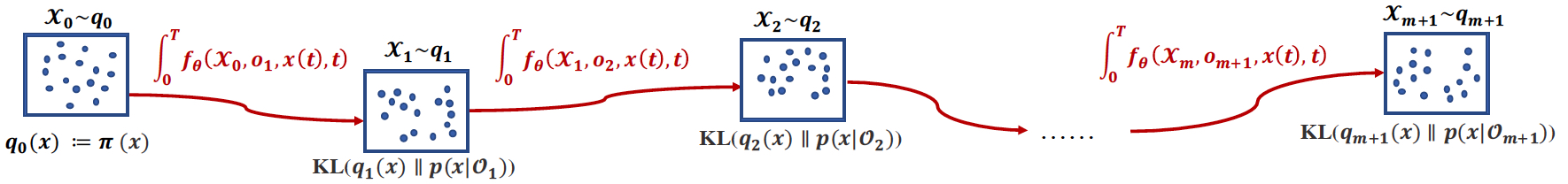}
    \vspace{-3mm}
    \caption{\small Sequential Bayesian inference as a deterministic flow of particles.}
    \label{fig:seq-framework}
    \vspace{-4mm}
\end{figure*}

\citet{chen2019particle} designed a continuous particle flow operator $\gF$ to realize the Bayes update, called {\bf meta particle flow} (MPF). In the MPF framework (Fig.~\ref{fig:seq-framework}), a prior distribution $\pi(\vx)$, or, the current posterior $\pi_m(\vx):=p(\vx|\gO_m)$ will be approximated by a set of $N$ {equally weighted} particles $\gX_m = \{\vx_m^1,\ldots,\vx_m^N\}$; 
and then, given an observation $o_{m+1}$, the flow operator $\gF(\gX_m, \vx_m^n, o_{m+1})$ will transport each particle $\vx_m^n$ to a new particle $\vx_{m+1}^n$ corresponding to the updated posterior $p(\vx|\gO_{m+1})$.

In a high-level, the MPF operator $\gF$ as a continuous deterministic flow, which propagates the locations of particles and the values of their probability density simultaneously through a dynamical system described by ordinary differential equations (ODEs). 

\subsection{Bayesian Inference as Particle Flow}

Starting with $N$ particles $\gX_0=\{\vx_0^1,\ldots,\vx_0^N\}$ sampled i.i.d. from a prior distribution $ \pi(\vx)$, given an observation $o_1$, the operator $\gF$ will transport the particles to $\gX_{1}=\{\vx_{1}^1,\ldots,\vx_{1}^N\}$ to estimate the posterior $p(\vx|\gO_1)\propto \pi(\vx)p(o_1|\vx)$. The transformation is modeled as the solution of an initial value problem of an ordinary differential equation (ODE). That is, $\forall n$,
\begin{equation*}
    \left\{\begin{tabular}{l}
        $\frac{d\vx}{dt} = f(\gX_0,o_{1},\vx(t),t),~\forall t\in(0,T]$ \cr
        $\vx(0) = \vx_0^n$
    \end{tabular}\right.
    \Longrightarrow {\text{gives}}~\vx_1^n =\vx(T),
\end{equation*}
where the flow velocity $f$ takes the observation $o_1$ as the input, and determines both the direction and the speed of the change of $\vx(t)$. In the above ODE model, each particle $\vx_0^n$ sampled from the prior gives an initial value $\vx(0)$, and then the flow velocity $f$ will evolve the particle continuously and deterministically. At the terminal time $T$, we will take the result of the evolution $\vx(T)$ as the transformed particle $\vx_1^n$ for estimating the posterior. 

Applying this ODE-based transformation sequentially as new observations $o_2, o_3,\ldots$ arrive, a recursive particle flow Bayes operator is defined as
\begin{align}
    \vx_{m+1}^n 
    & =\gF(\gX_m, o_{m+1}, \vx_m^n) \nonumber \\
    &:=\vx_m^n +\textstyle{ \int_{0}^{T}}f(\gX_m, o_{m+1}, \vx(t),t)\,dt. \label{eq:flow}
\end{align}

\subsubsection{Property of Continuous Deterministic  Flow}\label{sec:mass-transport}

The continuous transformation of $\vx(t)$ described by the differential equation $d\vx/dt = f$ defines a \emph{deterministic} flow for each particle. Let $q(\vx, t)$ be the probability density of the continuous random variable $\vx(t)$. The change of this density is also determined by $f$. More specifically, the evolution of the density follows the widely known {\bf continuity equation}~\citep{Batchelor2000kinematics}:
\begin{align}\label{eq:continuity}
    \frac{\partial q(\vx,t)}{\partial t} = -\nabla_x\cdot (qf),
\end{align}
where $\nabla_x\cdot$ is the divergence operator.
Continuity equation is the mathematical expression for the law of {\it local conservation of mass} - mass can neither be created nor destroyed, nor can it ''teleport'' from one place to another.

Given continuity equation, one can describe the change of log-density by another differential equation, for which we state it as Theorem~\ref{thm:change-of-log-density}. 

\begin{theorem}\label{thm:change-of-log-density} If $d\vx/dt =f$, then the change in log-density follows the differential equation
\begin{equation}
    \frac{d \log(q(\vx, t))}{dt} = -\nabla_x \cdot f.
\end{equation}
\end{theorem}

Note: $d/dt$ is  material derivative (total derivative) and $\partial/\partial t$ is partial derivative. $dq/dt$ defines the rate of change of $q$ in a given particle as it moves along its trajectory $\vx=\vx(t)$ in the flow, while $\partial q/\partial t$ means the rate of change of $q$ at a particular point $\vx$ that is fixed in the space.

With theorem~\ref{thm:change-of-log-density}, we can compute the log-density of the particles associated with the Bayes operator $\gF$ by integrating across $(0,T]$ for each $n$:
\begin{align}
    \label{eq:density_evolve}
   \log q_{m+1}(x_{m+1}^n) = \log q_m(x_m^n) - \textstyle{\int_{0}^{T}}\nabla_x \cdot f\,dt. 
\end{align}

\subsection{Existence of Shared Flow Velocity}\label{sec:existence}

Does a shared flow velocity $f$ exist for different Bayesian inference tasks involving different priors and observations? If it does, what is the form of this function? These questions are non-trivial even for simple Gaussian case. 

For instance, let the prior $\pi(x) = \gN(0, \sigma_x)$ and the likelihood $p(o|x) = \gN(x, \sigma)$ both be one dimensional Gaussian distributions. Given an observation $o=0$, the posterior distribution of $x$ is also a Gaussian distributed as $\gN(0, (\sigma \cdot \sigma_x)/(\sigma+\sigma_x))$. This means that the  dynamical system $d\vx/dt = f$ needs to push a zero mean Gaussian distribution with covariance $\sigma_x$ to another zero mean Gaussian distribution with covariance $(\sigma \cdot \sigma_x)/(\sigma+\sigma_x)$ for any $\sigma_x$. It is not clear whether such a shared flow velocity function $f$ exists and what is the form for it. 

To resolve the existence issue, we consider the Langevin dynamics, which is a \emph{stochastic} process 
\begin{align}
    \label{eq:Langevin}
    d\vx(t) = &\nabla_x\log p (\vx|\gO_{m})p (o_{m+1}|\vx)\,dt +\sqrt{2}d\vw(t), 
\end{align}
where $d\vw(t)$ is a standard Brownian motion. This stochastic flow is very different in nature comparing to the deterministic flow in \Secref{sec:mass-transport}, where a fixed location $\vx(0)$ will always end up as the same location $\vx(t)$. Nonetheless, we established their connection and proved the following theorem.

\begin{theorem}
There exists a fixed and deterministic flow velocity $f$ of the form
\begin{align}\label{eq:shared-f}
  \nabla_x\log p (\vx|\gO_{m})p (o_{m+1}|\vx)
  -w^*(p (\vx|\gO_{m}),p(o_{m+1}|\vx),t),
\end{align}
     such that its density $q(x,t)$ the same evolution as the density of Langevin dynamics.
\end{theorem}

\subsection{Parametrization}

In \Secref{sec:existence}, we introduce a shared flow velocity, which is in form of $f(p(\vx|\gO_m), p(o_{m+1}|\vx), \vx(t),t)$ as indicated by~\eqref{eq:shared-f}. We design the parameterization of $f$ based on this expression.

{(i) $p(\vx|\gO_m)\Rightarrow \gX_m$:}
In our particle flow framework, since we do not have full access to the density $p(\vx|\gO_{m})$ but have samples $\gX_m = \{\vx_m^1,\ldots,\vx_m^{N}\}$ from it, we can use these samples as surrogates for $p(\vx|\gO_{m})$. A related example is feature space embedding of distributions~\citep{smola2007hilbert}, 
$\mu_{\gX}(p):= \textstyle{\int_{\gX}}\phi(\vx)p(\vx)\,d\vx \approx \textstyle{\frac{1}{N}\sum_{n=1}^N}\phi(\vx^n),~\vx^n\sim p$. 

Ideally, if $\mu_{\gX}$ is an injective mapping from the space of probability measures over $\gX$ to the feature space, the resulting embedding can be treated as a sufficient statistic of the density and any information we need from $p(\vx|\gO_m)$ can be preserved. Hence, we represent $p(\vx|\gO_m)$ by $\frac{1}{N}\sum_{n=1}^N \phi(\vx_m^n)$, where $\phi(\cdot)$ is a nonlinear feature mapping to be learned. Since we use a neural version of $\phi(\cdot)$, 
this representation can also be regarded as a DeepSet~\citep{zaheer2017deep}.

{(ii) $p(o_{m+1}|\vx)\Rightarrow (o_{m+1}, \vx(t))$:}
In both Langevin dynamics and~\eqref{eq:shared-f}, the only term containing the likelihood is $\nabla_x \log p(o_{m+1}|\vx)$. Consequently, we can use this term as an input to $f$. In the case when the likelihood function is fixed, we can also simply use the observation $o_{m+1}$, which results in similar performance in our experiments.  

 Overall we will parameterize the flow velocity as 
\begin{align}\label{eq:interacting}
    f = \vh\left( \textstyle{\frac{1}{N}\sum_{n=1}^N} \phi(\vx_m^n),o_{m+1},\vx(t),t \right ),
\end{align}
where $\vh$ is a neural network and $\theta\in\Theta$ are parameters of $\vh$ and $\phi$. From now on, we will write $f = f_{\theta}(\gX_m, o_{m+1}, \vx(t),t)$, where $\theta\in\Theta$ is independent of $t$. In the next section, we will propose a meta learning framework for learning these parameters. 

% \begin{itemize}
%     \item we use a trick: $\pi_{\alpha}(\gX_m,o_{m+1}) = \{t_m, T_m\}$, the distance between different distribution pair can be different.
% \end{itemize}

\subsection{Meta Learning}

% \Le{Should have something explicit on task creation.} 
% \Le{Need to state more clearly why it is a multiple task learning setting.}

Since we want to learn a generalizable Bayesian inference operator, we need to create multiple inference tasks and design the corresponding training and testing algorithm. We will discuss these details below. 

% \begin{paragraph}
{\bf Multi-task Framework.}
We are interested in a training set $\gD_{train}$ containing multiple inference tasks.
Each task is a tuple
\begin{align*}
   \gT := \left(\pi(\vx), p(\cdot | \vx), \gO_M:=\{o_1,\ldots,o_M\}\right) \in \gD_{train}
\end{align*}
which consists of a prior distribution, a likelihood function and a sequence of $M$ observations. As we explained in previous sections, we want to learn a Bayesian operator $\gF$ that can be applied recursively to hit the targets $p(\vx|\gO_1), p(\vx|\gO_2),\cdots$ sequentially. Therefore, each task can also be interpreted as a sequence of $M$ sub-tasks:
 \begin{align*}
    \tau := (p(\vx|\gO_m), p(\cdot|\vx), o_{m+1}) \in (\pi(\vx), p(\cdot | \vx), \gO_M).
 \end{align*}
Therefore, each task is a sequential Bayesian inference and each sub-task corresponds to one step Bayesian update.
%  \end{paragraph}

% \begin{paragraph}
{\bf Cumulative Loss Function.}
For each sub-task we define a loss $\text{KL}(q_m(\vx)||p(\vx|\gO_{m+1}))$, where $q_m(\vx)$ is the distribution transported by $\gF$ at $m$-th stage and $p(\vx|\gO_{m+1})$ is the target posterior (see Fig.~\ref{fig:seq-framework} for illustration). Meanwhile, the loss for the corresponding sequential task will be $
 \textstyle{\sum_{m=1}^M}\text{KL}(q_m(\vx)||p(\vx|\gO_m) )
$, which sums up the sub-loss of all intermediate stages. Since its optimality is independent of normalizing constants of $p(\vx|\gO_m)$, it is equivalent to minimize the negative evidence lower bound (ELBO)
 \begin{align}\label{eq:elbo}
  \gL(\gT) = {\sum_{m=1}^M} \sum_{n=1}^N \left(\log q_m(\vx_m^n)-\log p(\vx_m^n,\gO_m)\right).
 \end{align}
 \vspace{-3mm}

 The above expression is an empirical estimation using particles $\vx_m^n$. In each iteration during training phase, we will randomly sample a task from $\gD_{train}$ and compute the gradient of the above cumulative loss function to update the parameters of our MPF operator $\gF$.

\subsection{Experiments}\label{sec:experiments}

Experiments on multivariate Gaussian model, hidden Markov model and Bayesian logistic regression are conducted to demonstrate the generalization ability of MPF as a Bayesian operator and also the posterior estimation performance of MPF as an inference method. Results can be found in~\cite{chen2019particle}. 

\section{Conclusion}
In this paper, we draw a comprehensive connection between neural architecture and ODE discretization, and between supervised learning and optimal control, and between PMP and SGD. Such connection enables us to design new models or algorithms for machine learning problems. More importantly, it opens new avenues to attack problems associated with machine learning. 

\newpage

\bibliographystyle{plainnat}

\begin{thebibliography}{41}
\providecommand{\natexlab}[1]{#1}
\providecommand{\url}[1]{\texttt{#1}}
\expandafter\ifx\csname urlstyle\endcsname\relax
  \providecommand{\doi}[1]{doi: #1}\else
  \providecommand{\doi}{doi: \begingroup \urlstyle{rm}\Url}\fi

\bibitem[Andrieu et~al.(2003)Andrieu, De~Freitas, Doucet, and
  Jordan]{andrieu2003introduction}
Christophe Andrieu, Nando De~Freitas, Arnaud Doucet, and Michael~I Jordan.
\newblock An introduction to mcmc for machine learning.
\newblock \emph{Machine learning}, 50\penalty0 (1-2):\penalty0 5--43, 2003.

\bibitem[Balakrishnan et~al.(2006)Balakrishnan, Madigan,
  et~al.]{balakrishnan2006one}
Suhrid Balakrishnan, David Madigan, et~al.
\newblock A one-pass sequential monte carlo method for bayesian analysis of
  massive datasets.
\newblock \emph{Bayesian Analysis}, 1\penalty0 (2):\penalty0 345--361, 2006.

\bibitem[Batchelor(2000)]{Batchelor2000kinematics}
G.K. Batchelor.
\newblock \emph{Kinematics of the Flow Field}, pages 71--130.
\newblock Cambridge Mathematical Library. Cambridge University Press, 2000.

\bibitem[Boltyanskii et~al.(1960)Boltyanskii, Gamkrelidze, and
  Pontryagin]{boltyanskii1960theory}
Vladimir~Grigor'evich Boltyanskii, Revaz~Valer'yanovich Gamkrelidze, and
  Lev~Semenovich Pontryagin.
\newblock The theory of optimal processes. i. the maximum principle.
\newblock Technical report, TRW SPACE TECHNOLOGY LABS LOS ANGELES CALIF, 1960.

\bibitem[Canini et~al.(2009)Canini, Shi, and Griffiths]{canini2009online}
Kevin Canini, Lei Shi, and Thomas Griffiths.
\newblock Online inference of topics with latent dirichlet allocation.
\newblock In \emph{Artificial Intelligence and Statistics}, pages 65--72, 2009.

\bibitem[Chang et~al.(2019)Chang, Chen, Haber, and
  Chi]{chang2018antisymmetricrnn}
Bo~Chang, Minmin Chen, Eldad Haber, and Ed~H. Chi.
\newblock Antisymmetric{RNN}: A dynamical system view on recurrent neural
  networks.
\newblock In \emph{International Conference on Learning Representations}, 2019.

\bibitem[Chen et~al.(2018)Chen, Rubanova, Bettencourt, and
  Duvenaud]{chen2018neural}
Tian~Qi Chen, Yulia Rubanova, Jesse Bettencourt, and David~K Duvenaud.
\newblock Neural ordinary differential equations.
\newblock In \emph{Advances in Neural Information Processing Systems}, pages
  6571--6583, 2018.

\bibitem[Chen et~al.(2019)Chen, Dai, and Song]{chen2019particle}
Xinshi Chen, Hanjun Dai, and Le~Song.
\newblock Particle flow bayes’ rule.
\newblock In \emph{International Conference on Machine Learning}, pages
  1022--1031, 2019.

\bibitem[Chopin et~al.(2013)Chopin, Jacob, and
  Papaspiliopoulos]{chopin2013smc2}
Nicolas Chopin, Pierre~E Jacob, and Omiros Papaspiliopoulos.
\newblock Smc2: an efficient algorithm for sequential analysis of state space
  models.
\newblock \emph{Journal of the Royal Statistical Society: Series B (Statistical
  Methodology)}, 75\penalty0 (3):\penalty0 397--426, 2013.

\bibitem[Daum and Huang(2003)]{daum2003curse}
Fred Daum and Jim Huang.
\newblock Curse of dimensionality and particle filters.
\newblock In \emph{2003 IEEE Aerospace Conference Proceedings (Cat. No.
  03TH8652)}, volume~4, pages 4\_1979--4\_1993. IEEE, 2003.

\bibitem[Dinh et~al.(2014)Dinh, Krueger, and Bengio]{dinh2014nice}
Laurent Dinh, David Krueger, and Yoshua Bengio.
\newblock Nice: Non-linear independent components estimation.
\newblock \emph{International Conference on Learning Representations Workshop},
  2014.

\bibitem[Dinh et~al.(2017)Dinh, Sohl-Dickstein, and Bengio]{dinh2016density}
Laurent Dinh, Jascha Sohl-Dickstein, and Samy Bengio.
\newblock Density estimation using real nvp.
\newblock In \emph{International Conference on Learning Representations}, 2017.

\bibitem[Doucet et~al.(2001)Doucet, De~Freitas, and
  Gordon]{doucet2001introduction}
Arnaud Doucet, Nando De~Freitas, and Neil Gordon.
\newblock An introduction to sequential monte carlo methods.
\newblock In \emph{Sequential Monte Carlo methods in practice}, pages 3--14.
  Springer, 2001.

\bibitem[Dreyfus(1964)]{dreyfus1964some}
Stuart~E Dreyfus.
\newblock Some types of optimal control of stochastic systems.
\newblock \emph{Journal of the Society for Industrial and Applied Mathematics,
  Series A: Control}, 2\penalty0 (1):\penalty0 120--134, 1964.

\bibitem[Freno and Carlberg(2019)]{freno2019machine}
Brian~A Freno and Kevin~T Carlberg.
\newblock Machine-learning error models for approximate solutions to
  parameterized systems of nonlinear equations.
\newblock \emph{Computer Methods in Applied Mechanics and Engineering},
  348:\penalty0 250--296, 2019.

\bibitem[Gomez et~al.(2017)Gomez, Ren, Urtasun, and
  Grosse]{gomez2017reversible}
Aidan~N Gomez, Mengye Ren, Raquel Urtasun, and Roger~B Grosse.
\newblock The reversible residual network: Backpropagation without storing
  activations.
\newblock In \emph{Advances in neural information processing systems}, pages
  2214--2224, 2017.

\bibitem[Grathwohl et~al.(2019)Grathwohl, Chen, Bettencourt, and
  Duvenaud]{grathwohl2018ffjord}
Will Grathwohl, Ricky T.~Q. Chen, Jesse Bettencourt, and David Duvenaud.
\newblock Ffjord: Free-form continuous dynamics for scalable reversible
  generative models.
\newblock In \emph{International Conference on Learning Representations}, 2019.

\bibitem[Grohs et~al.(2018)Grohs, Hornung, Jentzen, and
  Von~Wurstemberger]{grohs2018proof}
Philipp Grohs, Fabian Hornung, Arnulf Jentzen, and Philippe Von~Wurstemberger.
\newblock A proof that artificial neural networks overcome the curse of
  dimensionality in the numerical approximation of black-scholes partial
  differential equations.
\newblock \emph{arXiv preprint arXiv:1809.02362}, 2018.

\bibitem[Haber and Ruthotto(2017)]{haber2017stable}
Eldad Haber and Lars Ruthotto.
\newblock Stable architectures for deep neural networks.
\newblock \emph{Inverse Problems}, 34\penalty0 (1):\penalty0 014004, 2017.

\bibitem[Han et~al.(2018)Han, Jentzen, and Weinan]{han2018solving}
Jiequn Han, Arnulf Jentzen, and E~Weinan.
\newblock Solving high-dimensional partial differential equations using deep
  learning.
\newblock \emph{Proceedings of the National Academy of Sciences}, 115\penalty0
  (34):\penalty0 8505--8510, 2018.

\bibitem[He et~al.(2016)He, Zhang, Ren, and Sun]{he2016deep}
Kaiming He, Xiangyu Zhang, Shaoqing Ren, and Jian Sun.
\newblock Deep residual learning for image recognition.
\newblock In \emph{Proceedings of the IEEE conference on computer vision and
  pattern recognition}, pages 770--778, 2016.

\bibitem[Hoffman et~al.(2013)Hoffman, Blei, Wang, and
  Paisley]{hoffman2013stochastic}
Matthew~D Hoffman, David~M Blei, Chong Wang, and John Paisley.
\newblock Stochastic variational inference.
\newblock \emph{The Journal of Machine Learning Research}, 14\penalty0
  (1):\penalty0 1303--1347, 2013.

\bibitem[Hutzenthaler et~al.(2019)Hutzenthaler, Jentzen, Kruse, and
  Nguyen]{hutzenthaler2019proof}
Martin Hutzenthaler, Arnulf Jentzen, Thomas Kruse, and Tuan~Anh Nguyen.
\newblock A proof that rectified deep neural networks overcome the curse of
  dimensionality in the numerical approximation of semilinear heat equations.
\newblock \emph{arXiv preprint arXiv:1901.10854}, 2019.

\bibitem[Jentzen et~al.(2018)Jentzen, Salimova, and Welti]{jentzen2018proof}
Arnulf Jentzen, Diyora Salimova, and Timo Welti.
\newblock A proof that deep artificial neural networks overcome the curse of
  dimensionality in the numerical approximation of kolmogorov partial
  differential equations with constant diffusion and nonlinear drift
  coefficients.
\newblock \emph{arXiv preprint arXiv:1809.07321}, 2018.

\bibitem[Kingma and Dhariwal(2018)]{kingma2018glow}
Durk~P Kingma and Prafulla Dhariwal.
\newblock Glow: Generative flow with invertible 1x1 convolutions.
\newblock In \emph{Advances in Neural Information Processing Systems}, pages
  10215--10224, 2018.

\bibitem[Larsson et~al.(2017)Larsson, Maire, and
  Shakhnarovich]{larsson2017fractalnet}
Gustav Larsson, Michael Maire, and Gregory Shakhnarovich.
\newblock Fractalnet: Ultra-deep neural networks without residuals.
\newblock In \emph{International Conference on Learning Representations}, 2017.

\bibitem[Li and Hao(2018)]{li2018optimal}
Qianxiao Li and Shuji Hao.
\newblock An optimal control approach to deep learning and applications to
  discrete-weight neural networks.
\newblock In \emph{International Conference on Machine Learning}, pages
  2991--3000, 2018.

\bibitem[Li et~al.(2017)Li, Chen, Tai, and Weinan]{li2017maximum}
Qianxiao Li, Long Chen, Cheng Tai, and E~Weinan.
\newblock Maximum principle based algorithms for deep learning.
\newblock \emph{The Journal of Machine Learning Research}, 18\penalty0
  (1):\penalty0 5998--6026, 2017.

\bibitem[Liao and Poggio(2016)]{liao2016bridging}
Qianli Liao and Tomaso Poggio.
\newblock Bridging the gaps between residual learning, recurrent neural
  networks and visual cortex.
\newblock \emph{arXiv preprint arXiv:1604.03640}, 2016.

\bibitem[Lu et~al.(2018)Lu, Zhong, Li, and Dong]{lu2018beyond}
Yiping Lu, Aoxiao Zhong, Quanzheng Li, and Bin Dong.
\newblock Beyond finite layer neural networks: Bridging deep architectures and
  numerical differential equations.
\newblock In \emph{International Conference on Machine Learning}, pages
  3282--3291, 2018.

\bibitem[Pontryagin et~al.(1962)Pontryagin, Mishchenko, Boltyanskii, and
  Gamkrelidze]{pontryagin1962mathematical}
Lev~Semenovich Pontryagin, EF~Mishchenko, VG~Boltyanskii, and RV~Gamkrelidze.
\newblock The mathematical theory of optimal processes.
\newblock 1962.

\bibitem[Reisinger and Zhang(2019)]{reisinger2019rectified}
Christoph Reisinger and Yufei Zhang.
\newblock Rectified deep neural networks overcome the curse of dimensionality
  for nonsmooth value functions in zero-sum games of nonlinear stiff systems.
\newblock \emph{arXiv preprint arXiv:1903.06652}, 2019.

\bibitem[Ruthotto and Haber(2018)]{ruthotto2018deep}
Lars Ruthotto and Eldad Haber.
\newblock Deep neural networks motivated by partial differential equations.
\newblock \emph{arXiv preprint arXiv:1804.04272}, 2018.

\bibitem[Smola et~al.(2007)Smola, Gretton, Song, and
  Sch{\"o}lkopf]{smola2007hilbert}
Alex Smola, Arthur Gretton, Le~Song, and Bernhard Sch{\"o}lkopf.
\newblock A hilbert space embedding for distributions.
\newblock In \emph{International Conference on Algorithmic Learning Theory},
  pages 13--31. Springer, 2007.

\bibitem[Wainwright et~al.(2008)Wainwright, Jordan,
  et~al.]{wainwright2008graphical}
Martin~J Wainwright, Michael~I Jordan, et~al.
\newblock Graphical models, exponential families, and variational inference.
\newblock \emph{Foundations and Trends{\textregistered} in Machine Learning},
  1\penalty0 (1--2):\penalty0 1--305, 2008.

\bibitem[Weinan et~al.(2017)Weinan, Han, and Jentzen]{weinan2017deep}
E~Weinan, Jiequn Han, and Arnulf Jentzen.
\newblock Deep learning-based numerical methods for high-dimensional parabolic
  partial differential equations and backward stochastic differential
  equations.
\newblock \emph{Communications in Mathematics and Statistics}, 5\penalty0
  (4):\penalty0 349--380, 2017.

\bibitem[Weinan et~al.(2019)Weinan, Han, and Li]{weinan2019mean}
E~Weinan, Jiequn Han, and Qianxiao Li.
\newblock A mean-field optimal control formulation of deep learning.
\newblock \emph{Research in the Mathematical Sciences}, 6\penalty0
  (1):\penalty0 10, 2019.

\bibitem[Welling and Teh(2011)]{welling2011bayesian}
Max Welling and Yee~W Teh.
\newblock Bayesian learning via stochastic gradient langevin dynamics.
\newblock In \emph{Proceedings of the 28th international conference on machine
  learning (ICML-11)}, pages 681--688, 2011.

\bibitem[Zaheer et~al.(2017)Zaheer, Kottur, Ravanbakhsh, Poczos, Salakhutdinov,
  and Smola]{zaheer2017deep}
Manzil Zaheer, Satwik Kottur, Siamak Ravanbakhsh, Barnabas Poczos, Ruslan~R
  Salakhutdinov, and Alexander~J Smola.
\newblock Deep sets.
\newblock In \emph{Advances in neural information processing systems}, pages
  3391--3401, 2017.

\bibitem[Zhang et~al.(2018)Zhang, Weinan, and Wang]{zhang2018monge}
Linfeng Zhang, E~Weinan, and Lei Wang.
\newblock Monge-amp$\backslash$ere flow for generative modeling.
\newblock \emph{arXiv preprint arXiv:1809.10188}, 2018.

\bibitem[Zhang et~al.(2017)Zhang, Li, Change~Loy, and Lin]{zhang2017polynet}
Xingcheng Zhang, Zhizhong Li, Chen Change~Loy, and Dahua Lin.
\newblock Polynet: A pursuit of structural diversity in very deep networks.
\newblock In \emph{Proceedings of the IEEE Conference on Computer Vision and
  Pattern Recognition}, pages 718--726, 2017.

\end{thebibliography}

\end{document}